\documentclass[10pt,twocolumn,letterpaper]{article}

\usepackage{cvpr}      % To produce the REVIEW version
\usepackage[ruled,vlined]{algorithm2e}
\usepackage{booktabs}
\usepackage{multirow}
\usepackage{amssymb}   % for \checkmark
\usepackage{makecell}
\definecolor{cvprblue}{rgb}{0.21,0.49,0.74}
\usepackage[pagebackref,breaklinks,colorlinks,allcolors=cvprblue]{hyperref}

\def\confName{CVPR}
\def\confYear{2026}

\title{EVLF: Early Vision-Language Fusion for Generative Dataset Distillation}

\author{
Wenqi Cai$^1$ \quad Yawen Zou$^1$ \quad Guang Li$^2$ \quad Chunzhi Gu$^3$ \quad Chao Zhang$^1$\\
$^1$University of Toyama \quad
$^2$Hokkaido University \quad
$^3$University of Fukui \\
}
\begin{document}
\maketitle

\begin{abstract}
Dataset distillation (DD) aims to synthesize compact training sets that enable models to achieve high accuracy with significantly fewer samples. Recent diffusion-based DD methods commonly introduce semantic guidance through late-stage cross-attention, where textual prompts tend to dominate the generative process. Although this strategy enforces label relevance, it diminishes the contribution of visual latents, resulting in over-corrected samples that mirror prompt patterns rather than reflecting intrinsic visual features. To solve this problem, we introduce an Early Vision-Language Fusion (EVLF) method that aligns textual and visual embeddings at the transition between the encoder and the generative backbone. By incorporating a lightweight cross-attention module at this transition, the early representations simultaneously encode local textures and global semantic directions across the denoising process. Importantly, EVLF is plug-and-play and can be easily integrated into any diffusion-based dataset distillation pipeline with an encoder. It works across different denoiser architectures and sampling schedules without any task-specific modifications. Extensive experiments demonstrate that EVLF generates semantically faithful and visually coherent synthetic data, yielding consistent improvements in downstream classification accuracy across varied settings. Source code is available at \url{https://github.com/wenqi-cai297/earlyfusion-for-dd/}.
\end{abstract}

\section{Introduction}
The rapid expansion of dataset scale and model capacity has driven significant progress in machine learning, but has also intensified concerns regarding computational and storage efficiency~\cite{deng2009imagenet,lecun2015deep}. To alleviate these issues, model compression techniques such as pruning~\cite{liu2017learning, he2019filter, ding2019centripetal, sharma2022rapid} and quantization~\cite{wu2016quantized, chen2021towards, chauhan2023post, xu2023eq} have been widely explored to reduce redundancy and deployment costs. More recently, dataset distillation (DD) has emerged as a complementary paradigm that focuses on data instead of model size, condensing large training sets into compact synthetic subsets that retain critical learning signals and allow models to achieve competitive accuracy with orders of magnitude fewer samples~\cite{liu2025evolution}.

\begin{figure}[t]
  \centering
  \includegraphics[width=\linewidth]{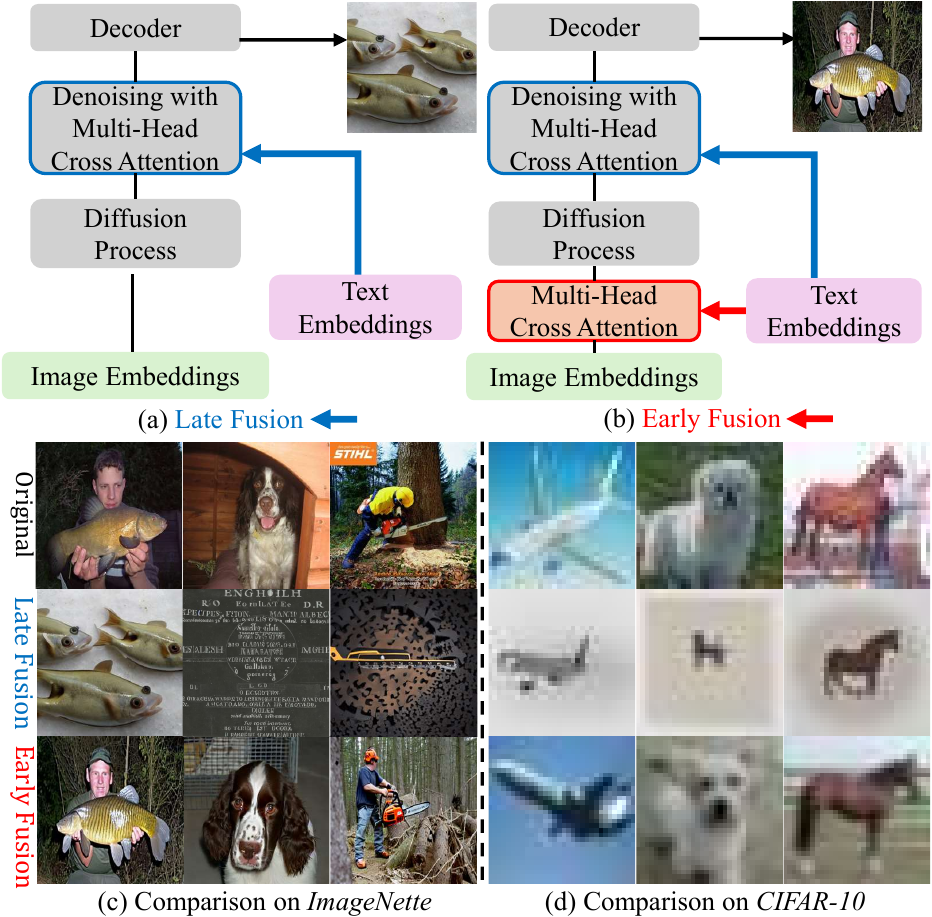}
  \caption{Comparison between traditional \textcolor{blue}{late-fusion} approaches and the proposed \textcolor{red}{EVLF}.
(a) Late-fusion methods inject textual prompts \emph{during} the denoising process, causing semantic signals to dominate visual latent representations.
(b) EVLF introduces vision-language alignment \emph{before} diffusion, allowing semantic cues and visual features to co-evolve throughout generation.
(c) Synthetic samples on ImageNette (256 $\times$ 256).
(d) Synthetic samples on CIFAR-10 (32 $\times$ 32).
Rows display real images, late-fusion results, and EVLF results. EVLF produces samples with stronger label fidelity and more coherent visual details.}
  \label{fig: illustration}
\end{figure}

Early DD methods were primarily based on meta-learning or data-matching objectives, but often exhibit substantial computational overhead or limited scalability when applied to high-resolution or large-scale datasets~\cite{wang2018dataset,nguyen2021dataset,lei2023comprehensive,zhao2020dataset,wang2022cafe}. Subsequent non-generative approaches have evolved in multiple directions, including meta-gradient optimization, gradient and trajectory matching, as well as distribution-level statistics alignment, to enhance both practicality and training stability~\cite{zhao2023improved,zhang2024m3d,yin2023squeeze,shao2024elucidating,he2024multisize}. Despite these advancements, achieving both high semantic fidelity and scalability remains challenging, motivating the emergence of generative diffusion-based DD as a promising new direction.

Recently, generative model-based DD has gained increasing attention due to its ability to produce diverse and high-resolution synthetic samples. Diffusion-based models have become the dominant backbone, with Latent Diffusion Models (LDMs)~\cite{rombach2022high} and Diffusion Transformers (DiTs)~\cite{peebles2023scalable} serving as representative architectures. Several approaches have extended diffusion synthesis for distillation. 
MinimaxDiffusion~\cite{gu2024efficient} formulates the distillation process as a minimax game within a DiT framework, enhancing both discriminability and representativeness. 
In parallel, D$^4$M~\cite{su2024d} augments LDMs through prototype-driven sampling by clustering latent embeddings and coupling them with label semantics. 
More recently, MGD$^3$~\cite{chan2025mgd} introduces a multimodal guidance mechanism that integrates seamlessly into the denoiser for both LDMs and DiTs, improving diversity and reducing redundancy in a plug-and-play manner.

Despite their success, diffusion-based DD methods inherit a core structural constraint from standard diffusion pipelines. In both LDMs~\cite{rombach2022high} and DiTs~\cite{peebles2023scalable}, semantic conditioning is injected after latent encoding and noise addition, and is applied during the denoising phase via cross-attention mechanisms inside the denoiser. This late-stage semantic injection amplifies prompt signals while weakening the influence of encoder-derived visual latents. As illustrated in Fig.~\ref{fig: illustration} (c)-(d), such late fusion methods often align strongly with label semantics but tend to compromise visual fidelity, resulting in unnatural shapes, text-like textures, and overly simplified object silhouettes. This behavior reveals a fundamental limitation of prompt-driven late fusion: because early latent representations contain mainly visual cues, semantics injected only during denoising act correctively rather than co-evolutionarily, pushing generation dynamics to overfit textual prompts and drift away from the encoder’s visual manifold. As a result, the model tends to produce samples that are label-relevant but visually distorted and lacking coherent structural detail.

In this work, we propose Early Vision-Language Fusion (EVLF) for dataset distillation. As illustrated in Fig.~\ref{fig: illustration} (a)-(b), instead of injecting semantics during denoising, EVLF performs vision-language alignment at the encoder-backbone interface, before the diffusion process begins. This produces latent representations that preserve encoder-derived visual structure while simultaneously encoding class-level semantic cues. The fusion module is trained to remain close to the original image latent and to align with the same-class text embeddings, ensuring that the resulting latent space reflects both visual fidelity and semantic relevance. Embedding semantics before denoising reshapes the generative trajectory: the denoiser now starts from an initialization that already integrates semantic and visual context, requires less prompt forcing, and operates closer to the underlying visual manifold. This mitigates the over-correction commonly observed in late fusion, where textual prompts dominate the denoising dynamics and distort structural details. 

Because EVLF is inserted only at the encoder-backbone handoff and does not depend on specific training schedules, it is a plug-and-play solution. It can be seamlessly integrated into any encoder-equipped diffusion-based DD pipeline. For pipelines that do not adapt the denoiser to the target dataset, an optional lightweight fine-tuning step can be applied to align noise prediction with fused representations. Compared to prior diffusion-based DD methods that introduce semantics exclusively during denoising and thus frequently experience fidelity degradation, EVLF consistently improves visual coherence and label alignment. Extensive experiments across diverse architectures, datasets, image-per-class (IPC) settings, and image resolutions demonstrate that EVLF provides robust and generalizable performance gains, surpassing state-of-the-art approaches while maintaining broad compatibility.

In summary, our contributions are as follows:
\begin{itemize}
    \item We identify a structural issue in diffusion-based dataset distillation: when semantics are injected only during denoising, prompt signals tend to dominate generation, causing over-correction and weakening the contribution of encoder-derived visual latents.
    \item We propose EVLF, which performs vision-language fusion \emph{before} denoising at the encoder-backbone interface. This produces latents that jointly encode visual structure and class semantics, guiding generation to remain close to the visual manifold.
    \item EVLF is plug-and-play and does not require modifying training schedules, loss functions, or denoiser architectures, making it directly compatible with a wide range of encoder-equipped diffusion-based DD pipelines.
    \item Extensive experiments across multiple datasets and IPC settings demonstrate that EVLF consistently improves semantic fidelity, visual coherence, diversity, and downstream classification accuracy over SOTA methods.
\end{itemize}

\section{Related Works}
Dataset distillation (DD) aims to synthesize compact yet informative datasets that preserve model performance while mitigating privacy risks associated with large real datasets~\cite{zhao2020dataset, zhao2021dataset, zhao2022synthesizing, zhao2023dataset, li2022awesome, sucholutsky2021secdd}. Early DD efforts focused on core-set selection~\cite{welling2009herding, chen2012super, rebuffi2017icarl, castro2018end}, which retains representative samples but limits flexibility in shaping data distributions. Optimization-based methods address this limitation and can be categorized into meta-learning and data-matching approaches. Meta-learning methods such as DD~\cite{wang2018dataset}, KIP~\cite{nguyen2021dataset}, RFAD~\cite{loo2022efficient}, and MDC~\cite{he2024multisize} formulate DD as a bi-level optimization problem, but incur high computational cost due to backpropagating through training trajectories. Data matching methods instead align model behavior under real and synthetic data, 
with examples including gradient matching (DC~\cite{zhao2020dataset}, DSA~\cite{zhao2021dataset}, IDM~\cite{zhao2023improved}) 
and trajectory alignment (MTT~\cite{cazenavette2022dataset}, APM~\cite{chen2023dataset}). Distribution-based approaches such as DM~\cite{zhao2023dataset}, CAFE~\cite{wang2022cafe}, and M3D~\cite{zhang2024m3d} match feature statistics to improve generality. However, these approaches often struggle to scale to high resolutions due to costly iterative optimization.
 
To improve scalability, recent work has explored decoupled distillation pipelines. Methods such as SRe$^2$L~\cite{yin2023squeeze}, G-VBSM~\cite{shao2024generalized}, RDED~\cite{sun2024diversity}, and EDC~\cite{shao2024elucidating} compress dataset statistics or leverage soft-label supervision to synthesize data more efficiently at high resolutions. While effective, these approaches rely on discriminative models and direct pixel- or feature-level optimization, which can limit semantic alignment and lead to less coherent textures or shapes. These drawbacks motivate the shift toward generative-model-based distillation, where synthesis is guided by generative priors rather than solely discriminative supervision.

Generative model-based DD has gained traction for producing diverse, high-resolution synthetic data~\cite{zhang2023dataset,gu2024efficient,su2024d,chan2025mgd,cazenavette2023generalizing,zhong2025hierarchical,moser2024latent}. Most methods adopt LDMs~\cite{rombach2022high} or DiTs~\cite{peebles2023scalable} for controllable synthesis. Specifically, MinimaxDiffusion~\cite{gu2024efficient} optimizes a minimax objective to encourage discriminability and representativeness, D$^4$M~\cite{su2024d} performs prototype-driven sampling to align clusters and semantics, MGD$^3$~\cite{chan2025mgd} introduces multimodal guidance to enhance diversity, and Zou et al.~\cite{zou2025dataset} introduce a vision-language distillation framework that leverages category-level textual prototypes alongside image prototypes to guide diffusion-based data generation. However, these approaches apply semantic conditioning during denoising, where the original latent contains only visual information. 
Consequently, textual prompts tend to dominate and over-correct the generation trajectory, producing samples that match labels but sacrifice structural fidelity and variation.
This limitation motivates early vision-language fusion, where semantics are introduced before noise injection, enabling balanced semantic and visual cues to co-exist throughout the diffusion process and improving both fidelity and diversity.

\begin{figure*}
    \centering
    \includegraphics[width=1\linewidth]{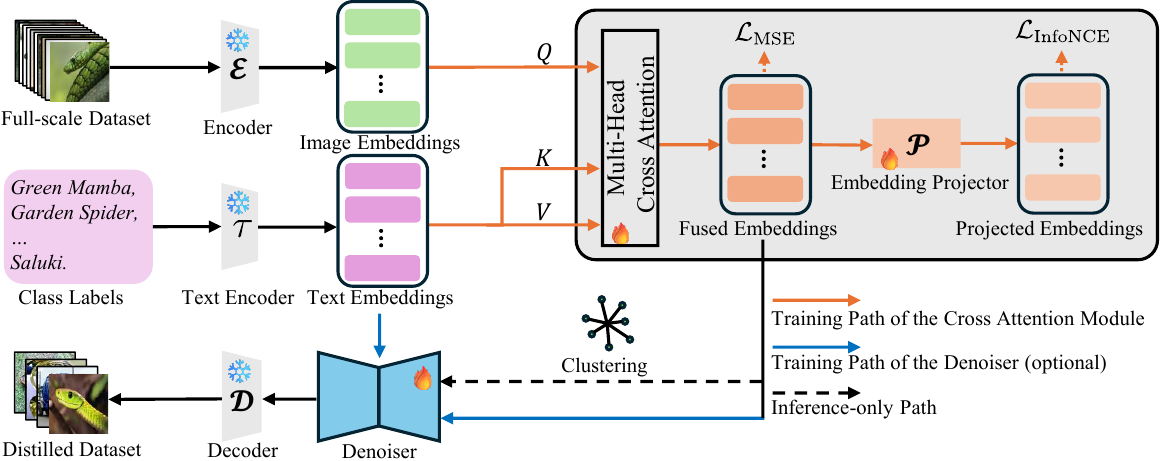}
    \caption{Overview of EVLF. Visual latents from a VAE and text embeddings from a text encoder are fused via cross-attention at the encoder-backbone interface. The fused embeddings are trained with $\mathcal{L}_{\mathrm{MSE}}$ for visual preservation and $\mathcal{L}_{\mathrm{InfoNCE}}$ for semantic alignment. Fused embeddings are clustered and decoded to produce the distilled synthetic dataset.}
    \label{fig: pipeline}
\end{figure*}

\section{Preliminaries}
\subsection{Dataset Distillation}
Dataset distillation aims to compress a large labeled dataset 
$T = \{(x_i, y_i)\}_{i=1}^{N_T}$ into a much smaller synthetic dataset 
$S = \{(\tilde{x}_i, \tilde{y}_i)\}_{i=1}^{N_S}$ with $N_S \ll N_T$, while maintaining comparable downstream performance.
Let $\text{Alg}(D, \theta_0)$ denote a learning algorithm that optimizes model parameters $\theta$ 
from initialization $\theta_0$ on dataset $D$ as:
\begin{equation}
\text{Alg}(D, \theta_0)
= \arg\min_{\theta} \ \mathbb{E}_{(x, y) \sim D} 
    \left[ \ell(x, y; \theta) \right],
\end{equation}
where $\ell(\cdot)$ denotes the task-specific loss function. 
The synthetic dataset $S$ is optimized such that a model trained only on $S$ generalizes well to real data from $T$:
\begin{equation}
\min_S \ \mathbb{E}_{(x, y) \sim T} 
    \left[ \ell(x, y; \theta_S^*) \right],
\quad \text{where } 
\theta_S^* = \text{Alg}(S, \theta_0).
\end{equation} The compression ratio is commonly controlled via an images per class (IPC) setting, which specifies the number of synthetic samples allocated to each class.

\subsection{Diffusion Models}
Diffusion models synthesize data by learning to reverse a fixed forward noising process.
Given a clean input $z_0 \sim q(z_0)$, the forward process gradually adds Gaussian noise to produce 
$\{z_t\}_{t=1}^T$:
\begin{equation}
    z_t = \sqrt{\alpha_t} z_0 + \sqrt{1 - \alpha_t}\,\epsilon, \quad \epsilon \sim \mathcal{N}(0, \mathbf{I}),
\end{equation}
where $\{\alpha_t\}$ is a predefined noise schedule. The reverse process trains a denoising network 
$\epsilon_\theta(z_t, t, c)$ to predict the added noise, optionally conditioned on auxiliary information $c$:
\begin{equation}
    \mathcal{L}_{\text{DM}} = \mathbb{E}_{z_0,\epsilon,t} \left[ \| \epsilon_\theta(z_t, t, c) - \epsilon \|_2^2 \right].
\end{equation}
Sampling begins from Gaussian noise $z_T$ and iteratively applies the learned denoiser to reconstruct $z_0$.

Two representative diffusion backbones are Latent Diffusion Models (LDMs)~\cite{rombach2022high} and Diffusion Transformers (DiTs)~\cite{peebles2023scalable}. 
LDMs operate in a compressed latent space using a VAE encoder-decoder and a U-Net with cross-attention conditioning, 
while DiTs replace the U-Net with a transformer-based denoiser to achieve better scalability.
 These architectures serve as the basis for our work. We investigate how semantic guidance can be injected earlier in the generative pipeline to better leverage encoder-derived visual latents before the denoising process begins.

\section{Method}
Our goal is to distill a large dataset into a compact synthetic set that preserves both semantic richness and visual fidelity. An overview of the proposed Early Vision-Language Fusion (EVLF) framework is shown in Fig.~\ref{fig: pipeline}. In standard diffusion-based distillation pipelines, textual semantics are injected during the denoising stage via cross-attention within the denoiser. This late-stage conditioning often causes textual prompts to dominate the generative trajectory, diminishing the contribution of encoder-derived visual information. In contrast, EVLF performs vision-language fusion immediately after encoding, before the diffusion process begins. Given an input image $x$ with label $y$, a VAE encoder produces a visual latent $z_{\text{img}} = \mathcal{E}(x)$, while a text encoder yields a class embedding $e_{\text{text}} = \mathcal{T}(y)$. We introduce a lightweight cross-attention module $\text{CA}$ to fuse the two: $z_{\text{fused}} = \text{CA
}(z_{\text{img}}, e_{\text{text}})$, and use $z_{\text{fused}}$ as the initial condition for the subsequent generative (diffusion) process. By anchoring semantics directly in the encoder latent space, EVLF ensures that textual cues guide but do not overwrite visual structure, thereby mitigating prompt dominance and preserving fine-grained visual characteristics throughout synthesis.

\subsection{Early Fusion Cross-Attention Module}
To ground semantic cues in the encoder-derived latent space, we integrate visual and textual information before any generative steps. Let the VAE encoder produce a spatial latent $z_{\text{img}}\in\mathbb{R}^{H\times W\times C}$ and the text encoder output a sequence of embeddings $e_{\text{text}}\in\mathbb{R}^{L\times C_t}$. Both representations are projected into a shared feature dimension $d$:
\begin{equation}
    \tilde{z}=\phi_{\text{img}}(z_{\text{img}})\in\mathbb{R}^{N\times d},\quad 
    \tilde{e}=\phi_{\text{text}}(e_{\text{text}})\in\mathbb{R}^{L\times d},
\end{equation}
where $N = H \times W$. Here, $\phi_{\text{img}}$ flattens the spatial latent into $N$ visual tokens and applies a linear projection, while $\phi_{\text{text}}$ linearly projects the text embeddings into the same feature space.

Cross-attention is performed using image tokens as queries and text tokens as keys and values:
\begin{equation}
Q=\tilde{z}W_Q,\qquad K=\tilde{e}W_K,\qquad V=\tilde{e}W_V,
\end{equation}
\begin{equation}
\text{Attn}(\tilde{z},\tilde{e})=\text{softmax}\!\left(\frac{QK^\top}{\sqrt{d}}\right)V.
\end{equation}
The attended features are merged with the visual tokens via a residual pathway, followed by layer normalization and a position-wise feed-forward transformation:
\begin{equation}
u=\text{LN}\!\big(\tilde{z}+\text{Attn}(\tilde{z},\tilde{e})\big),\quad 
z_{\text{fused}}=\psi(u)\in\mathbb{R}^{H\times W\times C},
\end{equation}
where $\psi$ restores the spatial arrangement and channel dimension.

The fused latent $z_{\text{fused}}$ is then forwarded to the subsequent generative process. By using image tokens as queries, semantics are grounded directly in the visual latent space, ensuring that textual cues guide rather than overwrite visual structure, thereby mitigating prompt-driven over-correction and preserving class-consistent appearance during synthesis.

\begin{algorithm}[t]
\caption{Training Process of EVLF}
\label{alg:efdd_compact}
1: \KwIn{Dataset $(X, Y)$: real images and prompts, $\mathcal{E}$: VAE encoder, 
$\mathcal{T}$: text encoder, 
CA: cross-attention module, $P$: projector, $\lambda_1, \lambda_2$: loss weights}
2: \KwOut{Trained cross-attention module CA}
3: \textit{/* Cross-Attention Training */} \\
4: \textbf{for} each batch $(x^i, y^i)$ in $(X,Y)$ \textbf{do} \\
5: \quad $z_{\text{img}}^i = \mathcal{E}(x^i)$ \\
6: \quad $e_{\text{text}}^i = \mathcal{T}(y^i)$ \\
7: \quad $z_{\text{fused}}^i = \text{CA}(z_{\text{img}}^i, e_{\text{text}}^i)$ \\
8: \quad $z_{\text{proj}}^i = P(z_{\text{fused}}^i)$ \\
9: \quad Compute $\mathcal{L}_{\text{MSE}}$ via Eq.~\ref{eq:mse}\\
10: \quad Compute $\mathcal{L}_{\text{InfoNCE}}$ via Eq.~\ref{eq:infonce}\\
11: \quad Update CA and $P$ via optimizing $\mathcal{L}_{\text{CA}}$ \\ 
\quad \quad \hspace{0.15cm} in Eq.~\ref{eq:cross_attention} \\
12: \textbf{end for} \\
\end{algorithm}
\subsection{Training the Cross-Attention Module}
The cross-attention module is trained with a dual-loss objective to preserve visual fidelity while enforcing semantic alignment. The first term encourages the fused latent $z_{\text{fused}}$ to remain close to the original image latent $z_{\text{img}}$, 
ensuring that text conditioning does not distort the underlying visual structure during early fusion.
\begin{equation}
\label{eq:mse}
    \mathcal{L}_{\text{MSE}} = \| z_{\text{fused}} - z_{\text{img}} \|_2^2.
\end{equation}

To incorporate semantics, we apply an InfoNCE loss that aligns $z_{\text{fused}}$ with class-level text embeddings. A learnable projector $P$ maps $z_{\text{fused}}$ into the same space as the text embeddings, giving $z_{\text{proj}} = P(z_{\text{fused}})$. For a batch of size $B$, let $M^{ij}$ denote whether samples $i$ and $j$ share the same class label:
\begin{equation}
M^{ij} =
\begin{cases}
1, & \text{if } y^i = y^j, \\
0, & \text{otherwise.}
\end{cases}
\end{equation}
With cosine similarity logits $s^{ij}$ computed between $z_{\text{proj}}^i$ and $e_{\text{text}}^j$, the InfoNCE term is:
\begin{equation}
\label{eq:infonce}
\mathcal{L}_{\text{InfoNCE}} = \frac{1}{B} \sum_{i=1}^{B} \left( -\log \frac{\sum_j M^{ij} \exp(s^{ij})}{\sum_j \exp(s^{ij})} \right).
\end{equation}
The final training objective is:
\begin{equation}
\label{eq:cross_attention}
    \mathcal{L}_{\text{CA}} = \lambda_1 \mathcal{L}_{\text{InfoNCE}} + \lambda_2 \mathcal{L}_{\text{MSE}},
\end{equation}
where $\lambda_1$ and $\lambda_2$ balance semantic consistency and visual preservation. Hyperparameter settings are described in Section~\ref{sec: hyperparameter}.

\subsection{Fine-tuning of the Denoiser}
Some distillation pipelines directly reuse a pretrained denoiser without adapting it to the target dataset. In such settings, the fused latent distribution introduced by EVLF may differ from the pretrained denoising prior. To address this, we optionally fine-tune the denoiser so that its noise prediction becomes consistent with both the target domain and the fused latent space. For example, when integrating EVLF into D$^4$M~\cite{su2024d}, we fine-tune the denoiser on fused representations, whereas for pipelines that already adapt or do not require adaptation, we keep it frozen.
The ablation results for this design choice are provided in Section~\ref{sec: ablation}.

Given a fused latent $z_{\text{fused}}$ and its corresponding text embedding $e_{\text{text}}$, the denoiser is trained using the standard diffusion objective:
\begin{equation}
    \mathcal{L}_{\text{DM}} = \mathbb{E}_{z_{\text{fused}}, \epsilon, t} \Big[ \big\| \epsilon_\theta(z_t, t, e_{\text{text}}) - \epsilon \big\|_2^2 \Big],
\end{equation}
where $z_t$ is the noised version of $z_{\text{fused}}$ at time step $t$, and $\epsilon \sim \mathcal{N}(0,\mathbf{I})$.

This step adds no extra modules and reuses the original diffusion objective. While optional, fine-tuning improves stability when the fused latent distribution differs from the pretrained backbone. Hyperparameter settings are provided in Section~\ref{sec: hyperparameter}.

\begin{table*}[t]
\centering
\caption{Dataset distillation results on ImageWoof across different IPC settings and test models. The best results in each row are in $\mathbf{bold}$, and the second-best are \underline{underlined}.}
\label{tab:imagewoof}
\resizebox{\textwidth}{!}{%
\begin{tabular}{c|l|cccccccccc|c}
\toprule
IPC (Ratio) & Test Model & Random & Herding & DiT & DM & IDC-1 & Minimax & D$^4$M & \textbf{D$^4$M+EVLF} & MGD$^3$ & \textbf{MGD$^3$+EVLF} & Full \\
\midrule
\multirow{3}{*}{10 (0.8\%)} & ConvNet-6   & $24.3 \! \pm \! {\scriptstyle 1.1}$ & $26.7 \! \pm \! {\scriptstyle 0.7}$ & $34.2 \! \pm \! {\scriptstyle 1.1}$ & $26.9 \! \pm \! {\scriptstyle 1.2}$ & $33.3 \! \pm \! {\scriptstyle 1.3}$ & $33.3 \! \pm \! {\scriptstyle 1.7}$ & $29.4 \! \pm \!                                      {\scriptstyle 0.9}$ & $\underline{34.3 \! \pm \! {\scriptstyle 2.4}}$ & $33.5 \! \pm \! {\scriptstyle 1.9}$ & $\mathbf{34.9 \! \pm \! {\scriptstyle 1.0}}$ & $86.4 \! \pm \! {\scriptstyle 0.2}$ \\
                            & ResNetAP-10 & $29.4 \! \pm \! {\scriptstyle 0.8}$ & $32.0 \! \pm \! {\scriptstyle 0.3}$ & $34.7 \! \pm \! {\scriptstyle 0.5}$ & $30.3 \! \pm \! {\scriptstyle 1.2}$ & $\underline{39.1 \! \pm \! {\scriptstyle 0.5}}$ & $36.2 \! \pm \! {\scriptstyle 3.2}$ & $33.2 \! \pm \! {\scriptstyle 2.1}$ & $37.3 \! \pm \! {\scriptstyle 0.7}$ & $36.6 \! \pm \! {\scriptstyle 0.9}$ & $\mathbf{39.3 \! \pm \! {\scriptstyle 0.3}}$ & $87.5 \! \pm \! {\scriptstyle 0.5}$ \\
                            & ResNet-18   & $27.7 \! \pm \! {\scriptstyle 0.9}$ & $30.2 \! \pm \! {\scriptstyle 1.2}$ & $34.7 \! \pm \! {\scriptstyle 0.4}$ & $33.4 \! \pm \! {\scriptstyle 0.7}$ & $\underline{37.3 \! \pm \! {\scriptstyle 0.2}}$ & $35.7 \! \pm \! {\scriptstyle 1.6}$ & $32.3 \! \pm \! {\scriptstyle 1.2}$ & $35.9 \! \pm \! {\scriptstyle 2.1}$ & $35.1 \! \pm \! {\scriptstyle 1.8}$ & $\mathbf{38.5 \! \pm \! {\scriptstyle 0.3}}$ & $89.3 \! \pm \! {\scriptstyle 1.2}$ \\
\midrule
\multirow{3}{*}{20 (1.6\%)} & ConvNet-6   & $29.1 \! \pm \! {\scriptstyle 0.7}$ & $29.5 \! \pm \! {\scriptstyle 0.3}$ & $36.1 \! \pm \! {\scriptstyle 0.8}$ & $29.9 \! \pm \! {\scriptstyle 1.0}$ & $35.5 \! \pm \! {\scriptstyle 0.8}$ & $37.3 \! \pm \! {\scriptstyle 0.1}$ & $34.0 \! \pm \!                                      {\scriptstyle 2.3}$ & $\underline{40.1 \! \pm \! {\scriptstyle 2.6}}$ & $36.2 \! \pm \! {\scriptstyle 1.6}$ & $\mathbf{40.2 \! \pm \! {\scriptstyle 0.5}}$ & $86.4 \! \pm \! {\scriptstyle 0.2}$ \\
                            & ResNetAP-10 & $32.7 \! \pm \! {\scriptstyle 0.4}$ & $34.9 \! \pm \! {\scriptstyle 0.1}$ & $41.1 \! \pm \! {\scriptstyle 0.8}$ & $35.2 \! \pm \! {\scriptstyle 0.6}$ & $43.4 \! \pm \! {\scriptstyle 0.3}$ & $43.3 \! \pm \! {\scriptstyle 2.7}$ & $40.1 \! \pm \! {\scriptstyle 1.6}$ & $42.8 \! \pm \! {\scriptstyle 0.2}$ & $\underline{44.5 \! \pm \! {\scriptstyle 2.8}}$ & $\mathbf{45.1 \! \pm \! {\scriptstyle 0.9}}$ & $87.5 \! \pm \! {\scriptstyle 0.5}$ \\
                            & ResNet-18   & $29.7 \! \pm \! {\scriptstyle 0.5}$ & $32.2 \! \pm \! {\scriptstyle 0.6}$ & $40.5 \! \pm \! {\scriptstyle 0.5}$ & $29.8 \! \pm \! {\scriptstyle 1.7}$ & $38.6 \! \pm \! {\scriptstyle 0.2}$ & $\underline{41.8 \! \pm \! {\scriptstyle 1.9}}$ & $38.4 \! \pm \! {\scriptstyle 1.1}$ & $40.7 \! \pm \! {\scriptstyle 1.3}$ & $40.3 \! \pm \! {\scriptstyle 2.5}$ & $\mathbf{42.1 \! \pm \! {\scriptstyle 0.3}}$ & $89.3 \! \pm \! {\scriptstyle 1.2}$ \\
\midrule
\multirow{3}{*}{50 (3.8\%)} & ConvNet-6   & $41.3 \! \pm \! {\scriptstyle 0.6}$ & $40.3 \! \pm \! {\scriptstyle 0.7}$ & $46.5 \! \pm \! {\scriptstyle 0.8}$ & $44.4 \! \pm \! {\scriptstyle 1.0}$ & $43.9 \! \pm \! {\scriptstyle 1.2}$ & $50.9 \! \pm \! {\scriptstyle 0.8}$ & $47.4 \! \pm \!                                      {\scriptstyle 0.9}$ & $\underline{52.5 \! \pm \! {\scriptstyle 0.9}}$ & $51.9 \! \pm \! {\scriptstyle 0.4}$ & $\mathbf{53.5 \! \pm \! {\scriptstyle 0.4}}$ & $86.4 \! \pm \! {\scriptstyle 0.2}$ \\
                            & ResNetAP-10 & $47.2 \! \pm \! {\scriptstyle 1.3}$ & $49.1 \! \pm \! {\scriptstyle 1.0}$ & $49.3 \! \pm \! {\scriptstyle 0.2}$ & $47.1 \! \pm \! {\scriptstyle 1.1}$ & $48.3 \! \pm \! {\scriptstyle 0.5}$ & $53.9 \! \pm \! {\scriptstyle 0.7}$ & $51.7 \! \pm \! {\scriptstyle 3.2}$ & $\underline{55.8 \! \pm \! {\scriptstyle 0.2}}$ & $55.6 \! \pm \! {\scriptstyle 1.0}$ & $\mathbf{59.0 \! \pm \! {\scriptstyle 1.1}}$ & $87.5 \! \pm \! {\scriptstyle 0.5}$ \\
                            & ResNet-18   & $47.9 \! \pm \! {\scriptstyle 1.8}$ & $48.3 \! \pm \! {\scriptstyle 1.2}$ & $50.1 \! \pm \! {\scriptstyle 0.5}$ & $46.2 \! \pm \! {\scriptstyle 0.6}$ & $48.3 \! \pm \! {\scriptstyle 0.8}$ & $53.7 \! \pm \! {\scriptstyle 0.6}$ & $53.7 \! \pm \! {\scriptstyle 2.2}$ & $\underline{58.1 \! \pm \! {\scriptstyle 0.9}}$ & $56.3 \! \pm \! {\scriptstyle 0.5}$ & $\mathbf{58.7 \! \pm \! {\scriptstyle 1.5}}$ & $89.3 \! \pm \! {\scriptstyle 1.2}$ \\
\midrule
\multirow{3}{*}{70 (5.4\%)} & ConvNet-6   & $46.3 \! \pm \! {\scriptstyle 0.6}$ & $46.2 \! \pm \! {\scriptstyle 0.6}$ & $50.1 \! \pm \! {\scriptstyle 1.2}$ & $47.5 \! \pm \! {\scriptstyle 0.8}$ & $48.9 \! \pm \! {\scriptstyle 0.7}$ & $51.3 \! \pm \! {\scriptstyle 0.6}$ & $50.5 \! \pm \!                                      {\scriptstyle 0.4}$ & $\underline{56.1 \! \pm \! {\scriptstyle 1.0}}$ & $53.1 \! \pm \! {\scriptstyle 0.9}$ & $\mathbf{56.7 \! \pm \! {\scriptstyle 1.3}}$ & $86.4 \! \pm \! {\scriptstyle 0.2}$ \\
                            & ResNetAP-10 & $50.8 \! \pm \! {\scriptstyle 0.6}$ & $53.4 \! \pm \! {\scriptstyle 0.9}$ & $54.3 \! \pm \! {\scriptstyle 0.9}$ & $51.7 \! \pm \! {\scriptstyle 0.9}$ & $52.8 \! \pm \! {\scriptstyle 1.8}$ & $57.0 \! \pm \! {\scriptstyle 0.2}$ & $54.7 \! \pm \! {\scriptstyle 1.6}$ & $\underline{59.6 \! \pm \! {\scriptstyle 1.2}}$ & $59.1 \! \pm \! {\scriptstyle 1.4}$ & $\mathbf{60.1 \! \pm \! {\scriptstyle 0.8}}$ & $87.5 \! \pm \! {\scriptstyle 0.5}$ \\
                            & ResNet-18   & $52.1 \! \pm \! {\scriptstyle 1.0}$ & $49.7 \! \pm \! {\scriptstyle 0.8}$ & $51.5 \! \pm \! {\scriptstyle 1.0}$ & $51.9 \! \pm \! {\scriptstyle 0.8}$ & $51.1 \! \pm \! {\scriptstyle 1.7}$ & $56.5 \! \pm \! {\scriptstyle 0.8}$ & $56.3 \! \pm \! {\scriptstyle 1.8}$ & $\underline{59.7 \! \pm \! {\scriptstyle 0.9}}$ & $59.1 \! \pm \! {\scriptstyle 0.1}$ & $\mathbf{60.5 \! \pm \! {\scriptstyle 0.8}}$ & $89.3 \! \pm \! {\scriptstyle 1.2}$ \\
\midrule
\multirow{3}{*}{100 (7.7\%)}& ConvNet-6   & $52.2 \! \pm \! {\scriptstyle 0.4}$ & $54.4 \! \pm \! {\scriptstyle 1.1}$ & $53.4 \! \pm \! {\scriptstyle 0.3}$ & $55.0 \! \pm \! {\scriptstyle 1.3}$ & $53.2 \! \pm \! {\scriptstyle 0.9}$ & $57.8 \! \pm \! {\scriptstyle 0.9}$ & $57.9 \! \pm \!                                      {\scriptstyle 1.5}$ & $\underline{60.0 \! \pm \! {\scriptstyle 0.2}}$ & $58.9 \! \pm \! {\scriptstyle 0.3}$ & $\mathbf{61.7 \! \pm \! {\scriptstyle 2.5}}$ & $86.4 \! \pm \! {\scriptstyle 0.2}$ \\
                            & ResNetAP-10 & $59.4 \! \pm \! {\scriptstyle 1.0}$ & $61.7 \! \pm \! {\scriptstyle 0.9}$ & $58.3 \! \pm \! {\scriptstyle 0.8}$ & $56.4 \! \pm \! {\scriptstyle 0.8}$ & $56.0 \! \pm \! {\scriptstyle 0.9}$ & $62.7 \! \pm \! {\scriptstyle 1.4}$ & $59.5 \! \pm \! {\scriptstyle 1.8}$ & $\underline{65.2 \! \pm \! {\scriptstyle 0.7}}$ & $64.3 \! \pm \! {\scriptstyle 1.5}$ & $\mathbf{68.1 \! \pm \! {\scriptstyle 0.9}}$ & $87.5 \! \pm \! {\scriptstyle 0.5}$ \\
                            & ResNet-18   & $61.5 \! \pm \! {\scriptstyle 1.3}$ & $59.3 \! \pm \! {\scriptstyle 0.7}$ & $58.9 \! \pm \! {\scriptstyle 1.3}$ & $60.2 \! \pm \! {\scriptstyle 1.0}$ & $58.3 \! \pm \! {\scriptstyle 1.2}$ & $62.7 \! \pm \! {\scriptstyle 0.4}$ & $63.8 \! \pm \! {\scriptstyle 1.3}$ & $\mathbf{67.8 \! \pm \! {\scriptstyle 1.9}}$ & $65.7 \! \pm \! {\scriptstyle 1.0}$ & $\underline{67.2 \! \pm \! {\scriptstyle 0.3}}$ & $89.3 \! \pm \! {\scriptstyle 1.2}$ \\
\bottomrule
\end{tabular}
}
\end{table*}

\begin{table}[t]
\centering
\caption{Comparison of SOTA methods under various IPC settings on ImageNette and ImageIDC. All results are on ResNetAP-10. Best in \textbf{bold}, second best \underline{underlined}.}
\label{tab:imagenette_idc}
\setlength{\tabcolsep}{3pt}
\renewcommand{\arraystretch}{1.05}
\resizebox{\linewidth}{!}{
\begin{tabular}{c c c c c c c c c c}
\toprule
\multicolumn{1}{c}{} & \multicolumn{1}{c}{\textbf{IPC}} & Random & DiT & DM & Minimax & D$^4$M & \textbf{D$^4$M+EVLF} & MGD$^3$ & \textbf{MGD$^3$+EVLF} \\
\midrule
\multirow{3}{*}{\textbf{Nette}} 
& 10 & $54.2 \! \pm \! {\scriptstyle 1.6}$ & $59.1 \! \pm \! {\scriptstyle 0.7}$ & $60.8 \! \pm \! {\scriptstyle 0.6}$ & $57.7 \! \pm \! {\scriptstyle 1.2}$ & $60.9 \! \pm \! {\scriptstyle 1.7}$ & $\underline{65.8 \! \pm \! {\scriptstyle 1.2}}$ & $64.3 \! \pm \! {\scriptstyle 1.0}$ & $\mathbf{66.0 \! \pm \! {\scriptstyle 1.6}}$ \\
& 20 & $63.5 \! \pm \! {\scriptstyle 0.5}$ & $64.8 \! \pm \! {\scriptstyle 1.2}$ & $66.5 \! \pm \! {\scriptstyle 1.1}$ & $64.7 \! \pm \! {\scriptstyle 0.8}$ & $66.3 \! \pm \! {\scriptstyle 1.3}$ & $\underline{71.7 \! \pm \! {\scriptstyle 0.5}}$ & $69.2 \! \pm \! {\scriptstyle 1.9}$ & $\mathbf{72.5 \! \pm \! {\scriptstyle 0.8}}$ \\
& 50 & $76.1 \! \pm \! {\scriptstyle 1.1}$ & $73.3 \! \pm \! {\scriptstyle 0.9}$ & $76.2 \! \pm \! {\scriptstyle 0.4}$ & $73.9 \! \pm \! {\scriptstyle 0.3}$ & $77.7 \! \pm \! {\scriptstyle 1.1}$ & $\mathbf{79.7 \! \pm \! {\scriptstyle 0.5}}$ & $79.2 \! \pm \! {\scriptstyle 1.9}$ & $\underline{79.5 \! \pm \! {\scriptstyle 0.4}}$ \\
\midrule
\multirow{3}{*}{\textbf{IDC}} 
& 10 & $48.1 \! \pm \! {\scriptstyle 0.8}$ & $54.1 \! \pm \! {\scriptstyle 0.4}$ & $52.8 \! \pm \! {\scriptstyle 0.5}$ & $51.9 \! \pm \! {\scriptstyle 1.4}$ & $47.7 \! \pm \! {\scriptstyle 0.5}$ & $\mathbf{57.3 \! \pm \! {\scriptstyle 1.5}}$ & $55.0 \! \pm \! {\scriptstyle 2.3}$ & $\underline{56.3 \! \pm \! {\scriptstyle 1.5}}$ \\
& 20 & $52.5 \! \pm \! {\scriptstyle 0.9}$ & $58.9 \! \pm \! {\scriptstyle 0.2}$ & $58.5 \! \pm \! {\scriptstyle 0.4}$ & $59.1 \! \pm \! {\scriptstyle 3.7}$ & $56.3 \! \pm \! {\scriptstyle 0.7}$ & $\underline{62.0 \! \pm \! {\scriptstyle 0.7}}$ & $61.7 \! \pm \! {\scriptstyle 1.0}$ & $\mathbf{64.1 \! \pm \! {\scriptstyle 0.3}}$ \\
& 50 & $68.1 \! \pm \! {\scriptstyle 0.7}$ & $64.3 \! \pm \! {\scriptstyle 0.6}$ & $69.1 \! \pm \! {\scriptstyle 0.8}$ & $69.4 \! \pm \! {\scriptstyle 1.4}$ & $67.8 \! \pm \! {\scriptstyle 1.0}$ & $\underline{72.1 \! \pm \! {\scriptstyle 0.3}}$ & $71.0 \! \pm \! {\scriptstyle 0.9}$ & $\mathbf{72.7 \! \pm \! {\scriptstyle 1.1}}$ \\
\bottomrule
\end{tabular}
}
\end{table}

\begin{table}[t]
\centering
\caption{Performance comparison on CIFAR-10 and CIFAR-100.}
\resizebox{\linewidth}{!}{%
\begin{tabular}{c|c|cccc}
\toprule
Dataset & IPC & SRe$^2$L & RDED & D$^4$M & \textbf{D$^4$M+EVLF} \\
\midrule
\multirow{2}{*}{CIFAR-10}  
        & 10 & $29.3 \!\pm\! 0.5$ & $37.1 \!\pm\! 0.3$ & $37.6 \!\pm\! 1.8$ & $\mathbf{45.7 \!\pm\! 0.5}$ \\
        & 50 & $45.0 \!\pm\! 0.7$ & $62.1 \!\pm\! 0.1$ & $71.7 \!\pm\! 1.2$ & $\mathbf{73.5 \!\pm\! 0.7}$ \\
\midrule
\multirow{2}{*}{CIFAR-100} 
        & 10 & $27.0 \!\pm\! 0.4$ & $42.6 \!\pm\! 0.2$ & $53.2 \!\pm\! 0.7$ & $\mathbf{56.2 \!\pm\! 0.4}$ \\
        & 50 & $50.2 \!\pm\! 0.4$ & $62.6 \!\pm\! 0.1$ & $66.0 \!\pm\! 0.2$ & $\mathbf{66.8 \!\pm\! 0.1}$ \\
\bottomrule
\end{tabular}}

\label{tab:cifar}
\end{table}

\begin{table}[t]
\centering
\caption{Performance comparison on Tiny-ImageNet.}
\resizebox{\linewidth}{!}{%
\begin{tabular}{c|c|cccc}
\toprule
Dataset & IPC & SRe$^2$L & RDED & D$^4$M & \textbf{D$^4$M+EVLF} \\
\midrule
\multirow{2}{*}{Tiny-ImageNet}  
            & 10 & $16.1 \!\pm\! 0.2$ & $41.9 \!\pm\! 0.2$ & $42.5 \!\pm\! 0.4$ & $\mathbf{49.2 \!\pm\! 0.4}$ \\
            & 50 & $41.1 \!\pm\! 0.4$ & $58.2 \!\pm\! 0.1$ & $55.8 \!\pm\! 0.1$ & $\mathbf{58.5 \!\pm\! 0.1}$ \\
\bottomrule
\end{tabular}}
\label{tab:tiny-imagenet}
\end{table}

% \begin{table}[t]
% \centering
% \caption{Performance comparison on ImageNet-1K.}
% \resizebox{\linewidth}{!}{%
% \begin{tabular}{c|c|cccccccc}
% \toprule
% Dataset & IPC & SRe$^2$L & RDED & DiT & Minimax & D$^4$M & \textbf{D$^4$M+EVLF} & MGD$^3$ & \textbf{MGD$^3$+EVLF} \\
% \midrule
% \multirow{2}{*}{ImageNet-1K} 
%             & 10 & $21.3 \!\pm\! 0.6$ & $42.0 \!\pm\! 0.1$ & $39.6 \!\pm\! 0.4$ & $44.3 \!\pm\! 0.5$ & $47.7 \!\pm\! 0.6$ & $48.3 \!\pm\! 0.3$ & $50.8 \!\pm\! 0.6$ & $\mathbf{51.3 \!\pm\! 0.3}$ \\
%             & 50 & $46.8 \!\pm\! 0.2$ & $56.5 \!\pm\! 0.1$ & $52.9 \!\pm\! 0.6$ & $58.6 \!\pm\! 0.3$ & $60.1 \!\pm\! 0.1$ & $60.6 \!\pm\! 0.0$ & $60.3 \!\pm\! 0.4$ & $\mathbf{61.9 \!\pm\! 0.1}$ \\
% \bottomrule
% \end{tabular}}
% \label{tab:imagenet_1k}
% \end{table}

\begin{table}[t]
\centering
\caption{Performance comparison on ImageNet-1K.}
\resizebox{\linewidth}{!}{%
\begin{tabular}{c|c|cccc}
\toprule
Dataset & IPC & \multicolumn{4}{c}{Accuracy (\%)} \\
\midrule
\multirow{8}{*}{\raisebox{-0.5\totalheight}{ImageNet-1K}} 
& \multirow{4}{*}{10} 
  & SRe$^2$L & RDED & DiT & Minimax \\
&   & $21.3 \!\pm\! 0.6$ & $42.0 \!\pm\! 0.1$ & $39.6 \!\pm\! 0.4$ & $44.3 \!\pm\! 0.5$ \\
&   & D$^4$M & \textbf{D$^4$M+EVLF} & MGD$^3$ & \textbf{MGD$^3$+EVLF} \\
&   & $47.7 \!\pm\! 0.6$ & $48.3 \!\pm\! 0.3$ & $50.8 \!\pm\! 0.6$ & $\mathbf{51.3 \!\pm\! 0.3}$ \\
\cmidrule(lr){2-6}
& \multirow{4}{*}{50} 
  & SRe$^2$L & RDED & DiT & Minimax \\
&   & $46.8 \!\pm\! 0.2$ & $56.5 \!\pm\! 0.1$ & $52.9 \!\pm\! 0.6$ & $58.6 \!\pm\! 0.3$ \\
&   & D$^4$M & \textbf{D$^4$M+EVLF} & MGD$^3$ & \textbf{MGD$^3$+EVLF} \\
&   & $60.1 \!\pm\! 0.1$ & $60.6 \!\pm\! 0.0$ & $60.3 \!\pm\! 0.4$ & $\mathbf{61.9 \!\pm\! 0.1}$ \\
\bottomrule
\end{tabular}}
\label{tab:imagenet_1k}
\end{table}

\section{Experiments}
\subsection{Datasets}
We evaluate our method on both small- and high-resolution benchmarks to assess its effectiveness across different data scales. For small-resolution settings, we use CIFAR-10 and CIFAR-100, each containing 60,000 natural images with 10 and 100 classes, respectively, which are standard testbeds for dataset distillation under low-resolution constraints. For high-resolution evaluation, we conduct experiments on ImageNet-1K and several of its commonly used subsets: ImageNette (10 easily separable classes), ImageWoof (fine-grained dog breeds), ImageIDC (domain-specific distribution shift), and Tiny-ImageNet (200 classes, 100K images) as a mid-scale benchmark. This selection enables a comprehensive comparison across varying resolutions, dataset scales, and task difficulties.

\subsection{Implementation Details}
\label{sec: hyperparameter}
The cross-attention module is trained for 4 epochs with a batch size of 16 using AdamW. We set the learning rate to $3\times10^{-4}$ for the cross-attention parameters and $1\times10^{-4}$ for the projector, with a weight decay of $1\times10^{-2}$ applied to both. To balance visual preservation and semantic alignment, we fix $\lambda_1 = 0.1$ for $\mathcal{L}_{\mathrm{InfoNCE}}$ and linearly increase $\lambda_2$ for $\mathcal{L}_{\mathrm{MSE}}$ from $0.05$ to $1.0$ over the first 2 training epochs.
After cross-attention training, the denoiser may be optionally fine-tuned on fused latents using the standard diffusion loss. In our experiments, D$^4$M employs this fine-tuning step, while MGD$^3$ retains the original denoiser.
We generate at $32 \times 32$ resolution for CIFAR-10/100, $256 \times 256$ for ImageNet-1K subsets, and $224 \times 224$ for full ImageNet-1K. For fair comparison, we follow the evaluation protocols of \cite{gu2024efficient} and \cite{sun2024diversity}. All experiments are performed on a single NVIDIA A5000 GPU.

\subsection{Comparison with SOTA Methods}  
We compare EVLF against two categories of dataset distillation methods.
The generative group includes MGD$^3$~\cite{chan2025mgd}, MinimaxDiffusion~\cite{gu2024efficient}, D$^4$M~\cite{su2024d}, and DiT-based diffusion backbones~\cite{gu2024efficient, peebles2023scalable}.
The non-generative group includes SRe$^2$L~\cite{yin2023squeeze}, RDED~\cite{sun2024diversity}, DM~\cite{zhao2023dataset}, IDC-1~\cite{kim2022dataset}, and Herding~\cite{welling2009herding}.
For fair comparison, we reproduced MGD$^3$, MinimaxDiffusion, and D$^4$M on ImageNette, ImageIDC, and ImageWoof using the authors’ official implementations. All reported results are averaged over three fixed seeds (0, 1, 2) and are presented as mean ± standard deviation.

\paragraph{ImageWoof.} 
We evaluate EVLF on ImageWoof under multiple IPC settings with ConvNet-6, ResNetAP-10, and ResNet-18 (Tab.~\ref{tab:imagewoof}). As a fine-grained dataset with high intra-class similarity, ImageWoof poses a challenging setting for distilled data synthesis. Across all architectures and IPC settings, EVLF consistently improves performance over the respective baselines, demonstrating its plug-and-play applicability and strong generalization.

At low IPC (e.g., IPC = 10), EVLF achieves 39.3\% accuracy on ResNetAP-10, outperforming the baseline by \textbf{2.7\%}. At higher IPC (e.g., IPC = 100), the improvement remains pronounced, surpassing MGD$^3$ by \textbf{3.8\%}. These results verify that EVLF effectively preserves fine-grained semantic cues and scales reliably across architectures and data regimes.

\paragraph{ImageNette and ImageIDC.} 
We examine EVLF on ImageNette and ImageIDC using ResNetAP-10 under IPC settings of 10, 20, and 50 (Tab.~\ref{tab:imagenette_idc}). Across all configurations, EVLF consistently outperforms the baselines. On ImageNette, EVLF achieves substantial gains, improving upon D$^4$M by an average of \textbf{4.9\%}.

ImageIDC presents a more challenging scenario due to its fine-grained categories; however, EVLF still delivers notable improvements. Under IPC = 10, EVLF surpasses D$^4$M by \textbf{9.6\%}, indicating strong robustness under limited sample budgets. These consistent gains confirm that early vision-language fusion effectively mitigates the over-correction issue in late-fusion pipelines, producing more semantically faithful and structurally coherent synthetic data that translates to stronger downstream performance.

\paragraph{CIFAR-10 and CIFAR-100.} 
We further evaluate EVLF on the low-resolution CIFAR-10 and CIFAR-100 datasets (32 $\times$ 32). As shown in Tab.~\ref{tab:cifar}, EVLF consistently outperforms previous state-of-the-art methods across IPC settings. Notably, at IPC = 10, EVLF surpasses D$^4$M by \textbf{8.1\%} on CIFAR-10. These results demonstrate that EVLF remains effective even under severe resolution and information constraints, indicating strong robustness and generalization across diverse visual conditions.

\paragraph{Tiny-ImageNet and ImageNet-1K.} 
We further validate EVLF on the mid-resolution Tiny-ImageNet dataset ($64 \times 64$) and the large-scale ImageNet-1K benchmark. As shown in Tab.~\ref{tab:tiny-imagenet}, EVLF consistently outperforms all competing methods on Tiny-ImageNet, yielding notable gains when integrated with D$^4$M. On ImageNet-1K (Tab.~\ref{tab:imagenet_1k}), EVLF continues to improve baseline performance and surpasses prior state-of-the-art approaches. These results demonstrate that the proposed early fusion strategy scales effectively from mid-resolution to full large-scale settings, providing stable improvements across dataset sizes and complexity levels.

\begin{table}[t]
\caption{Transfer learning results on target datasets using models pretrained on the distilled ImageNet-1K dataset.}
\centering
\small
\resizebox{\linewidth}{!}{
\begin{tabular}{lccccccc}
\toprule
Method & CIFAR-10 & CIFAR-100 & Dogs & Flowers \\
\midrule
w/o pre   & $88.66 \!\pm\! 0.09$ & $66.62 \!\pm\! 0.32$ & $24.59 \!\pm\! 0.46$ & $59.39 \!\pm\! 0.29$ \\
Random    & $88.46 \!\pm\! 0.09$ & $65.97 \!\pm\! 0.08$ & $23.08 \!\pm\! 0.40$ & $56.81 \!\pm\! 0.40$ \\
FRePo     & $87.88 \!\pm\! 0.20$ & $65.23 \!\pm\! 0.47$ & $22.05 \!\pm\! 0.45$ & $52.50 \!\pm\! 0.51$ \\
KRR-ST    & $89.33 \!\pm\! 0.19$ & $68.04 \!\pm\! 0.22$ & $35.51 \!\pm\! 0.45$ & $70.45 \!\pm\! 0.34$ \\
\midrule
\textbf{MGD$^3$+EVLF}   & $\mathbf{90.23 \!\pm\! 0.15}$ & $\mathbf{69.21 \!\pm\! 0.07}$ & $\mathbf{36.18 \!\pm\! 0.37}$ & $\mathbf{76.03 \!\pm\! 0.53}$ \\
\bottomrule
\end{tabular}
\label{tab: transfer_imagenet}
}
\end{table}

\paragraph{Transfer Learning.}
To further evaluate the generalization capability of the distilled datasets, we conduct transfer learning experiments following the protocol of KRR-ST~\cite{lee2024self}. Specifically, a ConvNet-4 model is first pre-trained on the distilled ImageNet-1K subset (0.8× of the original training size) and then fine-tuned on downstream target datasets. As shown in Tab.~\ref{tab: transfer_imagenet}, datasets distilled by MGD$^3$+EVLF yield consistently higher fine-tuning accuracy than those distilled by prior methods. This indicates that our synthesized data better preserves discriminative semantics and class-consistent visual structure, enabling more effective feature transfer across tasks.

\subsection{Ablation Studies}
\label{sec: ablation}
\paragraph{Impact of Denoiser Fine-Tuning and CrossAttention.}
We compare four variants of the D$^4$M pipeline:
\begin{enumerate}[label=(\arabic*)]
    \item \textbf{Baseline}: D$^4$M with a pretrained denoiser.
    \item \textbf{D$^4$M + Denoiser Fine-Tuning}: The denoiser is fine-tuned on original visual embeddings from ImageIDC.
    \item \textbf{D$^4$M + CrossAttention}: CrossAttention module is applied while keeping the denoiser frozen.
    \item \textbf{D$^4$M + CrossAttention + Denoiser Fine-Tuning}: The denoiser is fine-tuned on fused embeddings.
\end{enumerate}
We evaluate all variants using ResNetAP-10 under IPC settings of 10, 20, and 50.

As shown in Tab.~\ref{tab: ablation}, both CrossAttention and denoiser fine-tuning independently improve performance over the baseline, and their combination yields the best results across all IPC settings. This confirms that early fusion addresses semantic over-correction, while denoiser adaptation helps align the generative prior with the fused latent distribution.

\begin{table}[t]
\centering
\caption{Ablation on the contributions of denoiser fine-tuning (FT.) and the CrossAttention module (CA.) within the D$^4$M pipeline. Results are reported on ImageIDC with ResNetAP-10 under varying IPC settings.}
\label{tab: ablation}
\setlength{\tabcolsep}{4pt}
\renewcommand{\arraystretch}{1.05}
\footnotesize

\resizebox{\columnwidth}{!}{%
\begin{tabular}{cccccc}
\toprule
\multirow{2}{*}{Method} & \multicolumn{2}{c}{Ablation Components} & \multicolumn{3}{c}{IPC} \\
\cmidrule(lr){2-3} \cmidrule(lr){4-6}
& FT. & CA. & 10 & 20 & 50 \\
\midrule
D$^4$M & -- & -- & $47.7 \!\pm\! 0.5$ & $56.3 \!\pm\! 0.7$  &  $67.8 \!\pm\! 1.0$ \\
\midrule
\textbf{D$^4$M+EVLF} &
\makecell[c]{$\checkmark$\\--\\$\checkmark$} &
\makecell[c]{--\\$\checkmark$\\$\checkmark$} &
\makecell[c]{$54.1 \!\pm\! 0.7$\\$51.1 \!\pm\! 2.5$\\$\mathbf{57.3 \!\pm\! 1.5}$} &
\makecell[c]{$61.1 \!\pm\! 0.1$\\$57.5 \!\pm\! 0.3$\\$\mathbf{62.0 \!\pm\! 0.7}$} &
\makecell[c]{$70.3 \!\pm\! 0.9$\\$69.1 \!\pm\! 2.3$\\$\mathbf{72.1 \!\pm\! 0.3}$} \\
\bottomrule
\end{tabular}%
}
\end{table}

\begin{figure}[t]
  \centering
  \includegraphics[width=\linewidth]{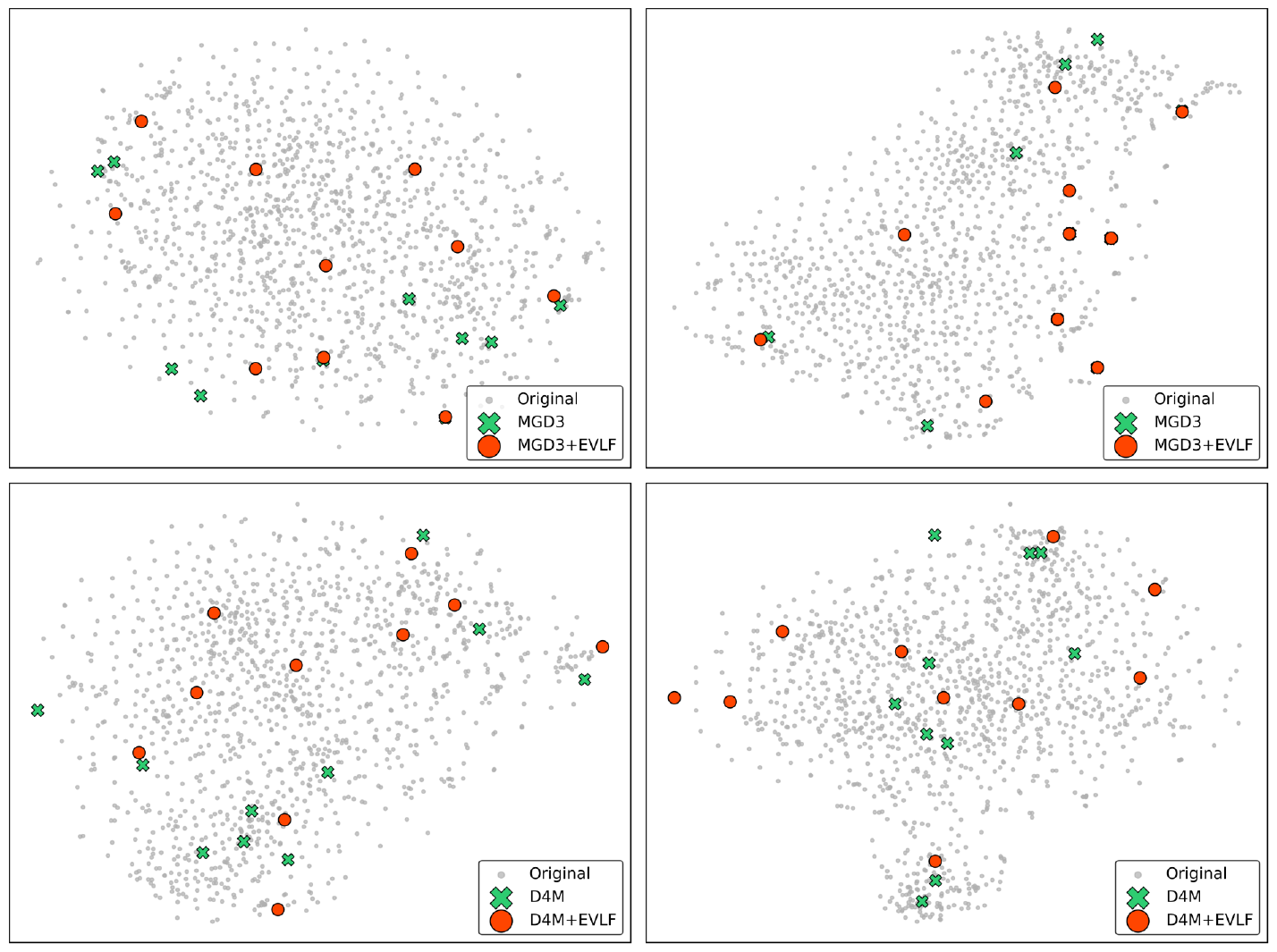}
  \caption{t-SNE visualization of synthetic and real samples on four ImageNet-1K classes. D$^4$M~\cite{su2024d} and MGD$^3$~\cite{chan2025mgd} produce synthetic samples that occupy limited regions of the real-data manifold. With EVLF, the synthesized samples cover a broader and more varied region, indicating improved diversity and distributional alignment.
}
  \label{fig: t-SNE}
\end{figure}

\paragraph{t-SNE Visualization.} Prior work has shown that the diversity of distilled samples strongly correlates with downstream performance~\cite{bo2025understanding, chan2025mgd}. To examine distributional coverage, we visualize the t-SNE embeddings of synthetic datasets generated by D$^4$M, MGD$^3$, and their respective variants augmented with EVLF. As shown in Fig.~\ref{fig: t-SNE}, D$^4$M and MGD$^3$ produce embeddings that occupy relatively narrow regions of the real-data manifold, indicating limited diversity. In contrast, incorporating EVLF yields embeddings that span a broader portion of the manifold, suggesting improved intra-class variation and richer class-wise representation coverage.

\paragraph{Parameter Analysis.}
We analyze the effect of the text-injection weight $\lambda_{1}$ in $\mathcal{L}_{\mathrm{InfoNCE}}$ from two perspectives: validation accuracy and distributional coverage. Coverage is measured following~\cite{naeem2020reliable}.
For each real sample, we compute the distance to its 20th nearest real neighbor to define a local radius.
A generated sample is considered covered if it falls within the radius of any real point, and the coverage score is the proportion of generated samples that satisfy this condition.

\begin{figure}[t]
  \centering
  \includegraphics[width=\linewidth]{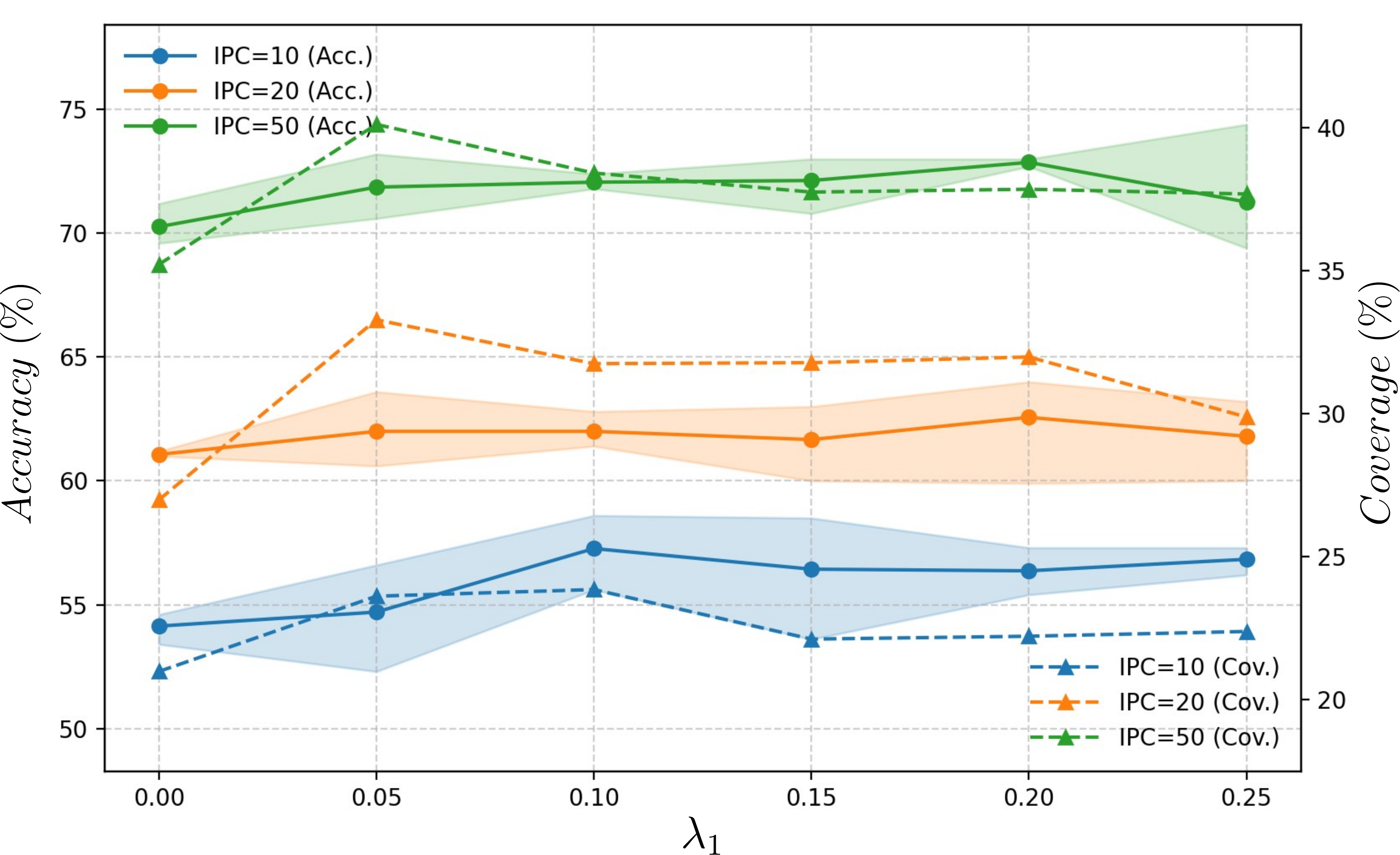}
  \caption{Parameter Analysis of $\lambda_1$ on ImageIDC.}
  \label{fig: linechart}
\end{figure}

As shown in Fig.~\ref{fig: linechart}, enabling text injection ($\lambda_{1} > 0$) leads to notable gains in both accuracy and coverage, while $\lambda_{1}=0$, i.e., no EVLF, results in over-corrected generations dominated by late-stage prompt conditioning, producing visually repetitive samples with reduced fidelity. Once EVLF is introduced, coverage increases substantially, indicating greater visual diversity and more faithful alignment with the real-data manifold. Further increasing $\lambda_{1}$ causes only minor fluctuations in both metrics, suggesting that EVLF is robust and not highly sensitive to this parameter. We adopt $\lambda_{1}=0.10$ as the default setting, as it yields the most stable accuracy (lowest variance in the shaded confidence region).

\subsection{Visualization}
\begin{figure}[t]
  \centering
  \includegraphics[width=\linewidth]{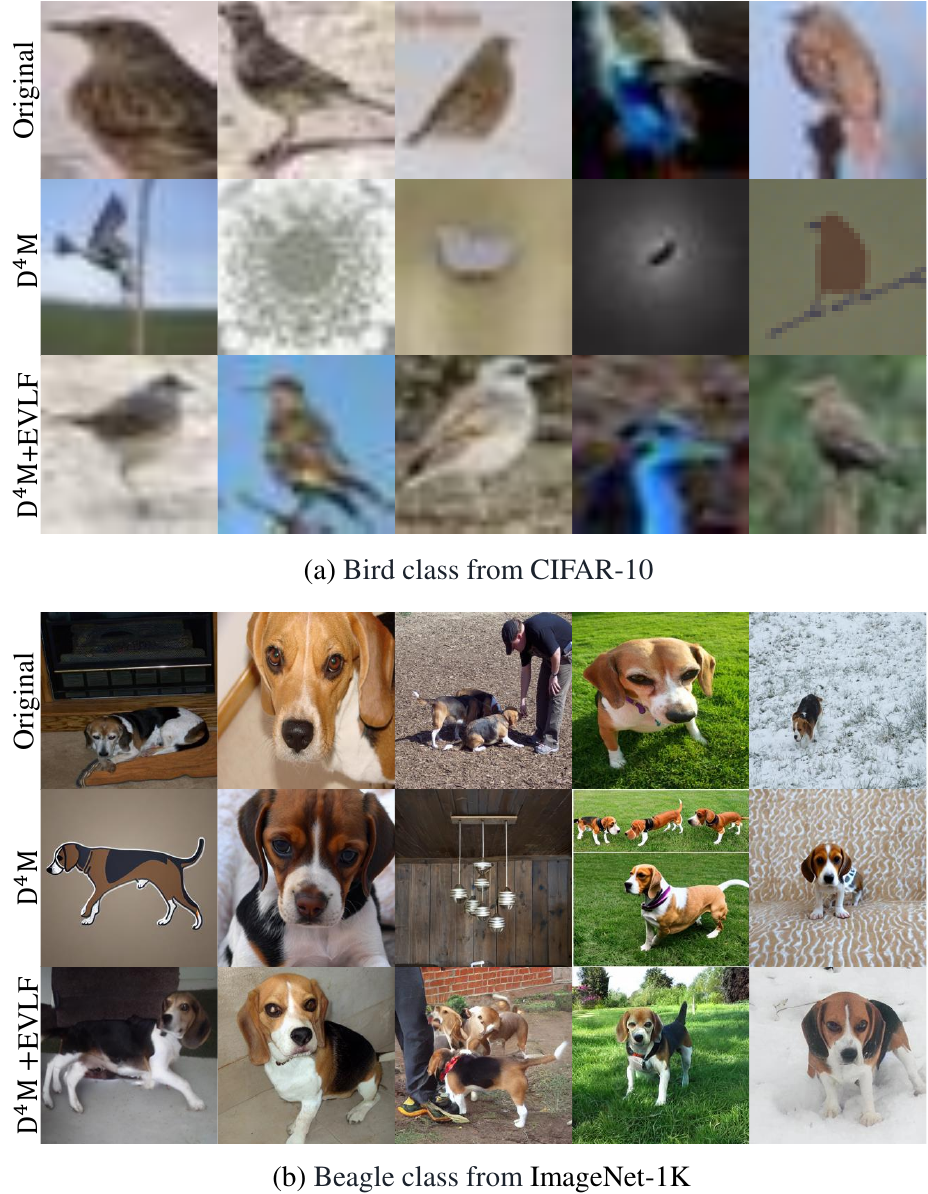}
  \caption{Visualization of synthesized images generated by D$^4$M and our EVLF under low- and high-resolution settings. (a) Bird class from CIFAR-10, and (b) Beagle class from ImageNet-1K. EVLF produces samples with clearer structure, richer textures, and improved consistency with class semantics across different image scales.}
  \label{fig: visualization}
\end{figure}

To further assess synthesis quality, we compare EVLF with prior methods at both low and high resolutions, as illustrated in Fig.~\ref{fig: visualization}.
Fig.~\ref{fig: visualization} (a) shows results for the Bird class on CIFAR-10. D$^4$M primarily captures coarse bird-like silhouettes with limited structural detail, whereas EVLF produces more natural and coherent shapes with clearer textures and greater intra-class variation.
Fig.~\ref{fig: visualization} (b) presents results for the Beagle class on ImageWoof. D$^4$M occasionally generates cartoonish or off-class artifacts, while EVLF yields richer texture patterns and visually consistent backgrounds.
These visual comparisons demonstrate that EVLF effectively preserves label semantics while also maintaining visual fidelity and diversity, resulting in broader and more realistic coverage of the underlying data distribution.

\section{Conclusion}
We introduced Early Vision-Language Fusion (EVLF), a plug-and-play method that integrates textual semantics into the visual latent space before denoising through a lightweight cross-attention mechanism. By grounding semantic cues at the encoder stage, EVLF mitigates prompt-induced over-correction and enables the generation of synthetic datasets that are both semantically faithful and visually coherent. The approach is architecture-agnostic and can be seamlessly incorporated into existing diffusion-based distillation pipelines without modifying their training objectives or model structures.
\paragraph{Limitations and Future Works}
Our current formulation focuses on class-level conditioning and does not address instance-level or multi-label scenarios. Future work will explore extending EVLF to instance-aware and compositional prompts to further enhance fine-grained control and sample diversity while preserving semantic consistency.

{
    \small
    \bibliographystyle{ieeenat_fullname}
    \bibliography{main}
}
\end{document}